\documentclass[conference]{IEEEtran}
\IEEEoverridecommandlockouts
\usepackage{cite}
\usepackage{amsmath,amssymb,amsfonts}
\usepackage{algorithmic}
\usepackage{graphicx}
\usepackage{textcomp}
\usepackage{xcolor}
\usepackage{dsfont}
\def\BibTeX{{\rm B\kern-.05em{\sc i\kern-.025em b}\kern-.08em
    T\kern-.1667em\lower.7ex\hbox{E}\kern-.125emX}}
\begin{document}

\title{Robust and Scalable Hyperdimensional Computing With Brain-Like Neural Adaptations
\vspace{-4mm}

}

\author{Junyao Wang$^*$ (\textit{junyaow4@uci.edu}, Graduate)\\
Advisor: Mohammad Abdullah Al Faruque (\textit{alfaruqu@uci.edu})\\
\textit{Department of Computer Science, University of California, Irvine, United States, 92697}
\vspace{-4mm}}

\maketitle
\let\thefootnote\relax\footnotetext{*Manuscript for the Student Research Competition in the 42$^{\text{nd}}$ International Conference on Computer-Aided Design (ICCAD) presented by the author.}
\section{Problem and Motivation}
The Internet of Things (IoT) has facilitated many applications utilizing edge-based machine learning (ML) methods to analyze locally collected data.
Unfortunately, popular ML algorithms often require intensive computations beyond the capabilities of today's IoT devices. 
Brain-inspired hyperdimensional computing (HDC) has been introduced to address this issue. 
However, existing HDCs use static encoders, requiring extremely high dimensionality and hundreds of training iterations to achieve reasonable accuracy. 
This results in a huge efficiency loss, severely impeding the application of HDCs in IoT systems. 
We observed that a  main cause is that the encoding module of existing HDCs lacks the capability to utilize and adapt to information learned during training.
In contrast, as shown in Fig.\ref{fig: intro}(a), neurons in human brains dynamically regenerate all the time and provide more useful functionalities when learning new information~\cite{andersen2003aging}. 
While the goal of HDC is to exploit the high-dimensionality of randomly generated base hypervectors to represent the information as a pattern of neural activity, it remains challenging for existing HDCs to support a similar behavior as brain neural regeneration. 
In this work, we present \textbf{\textit{dynamic}} HDC learning frameworks that identify and regenerate undesired dimensions to provide adequate accuracy with significantly lowered dimensionalities, thereby accelerating both the training and inference.

\vspace{-1mm}
\section{Background and Related Work} \label{sec:background}
\subsection{IoT and Edge-based Learning}
Many novel frameworks and libraries have been developed to customize popular ML algorithms on resource-constrained computing platforms, including TinyML~\cite{warden2019tinyml}, TensorFlow Lite~\cite{david2021tensorflow}, edge-ml~\cite{sakr2020machine}, X-Cube-AI~\cite{xcubeai}, etc. 
However, these learning methods often require large amounts of training samples and multiple training cycles beyond the capabilities of today's IoT devices.  
Meanwhile, leveraging the learning structures and properties of target platforms, researchers have proposed a number of techniques to improve the efficiency of edge-based learning, e.g., split computing~\cite{ko2018edge}, federated learning~\cite{bonawitz2017practical, li2020federated}, knowledge distillation~\cite{luo2022keepedge}. 
These techniques are orthogonal to our method and can potentially be integrated with our approach for further enhanced learning performance.

\subsection{Hyperdimensional Computing}
 Prior studies have exhibited enormous success in various applications of HDCs~\cite{poduval2022graphd, burrello2019laelaps, kim2018efficient}. 
 However, existing HDCs use pre-generated static encoders 
 and thus require extremely high dimensionality to achieve reasonable accuracy~\cite{rahimi2007random}. 
To be best of our knowledge, we propose brain-like neural adaptation for HDC for the first time. 
We not only compress HDC models by eliminating dimensions playing minor roles in classification tasks, but also regenerate misleading or biased dimensions to improve model performance. 
Additionally, we fully optimize our learning framework with highly parallel matrix operations on high-dimensional space, and provide resource-efficient and hardware-friendly solutions for edge-based ML applications.
\begin{figure}[!t]
\centering
\includegraphics[width=\linewidth]{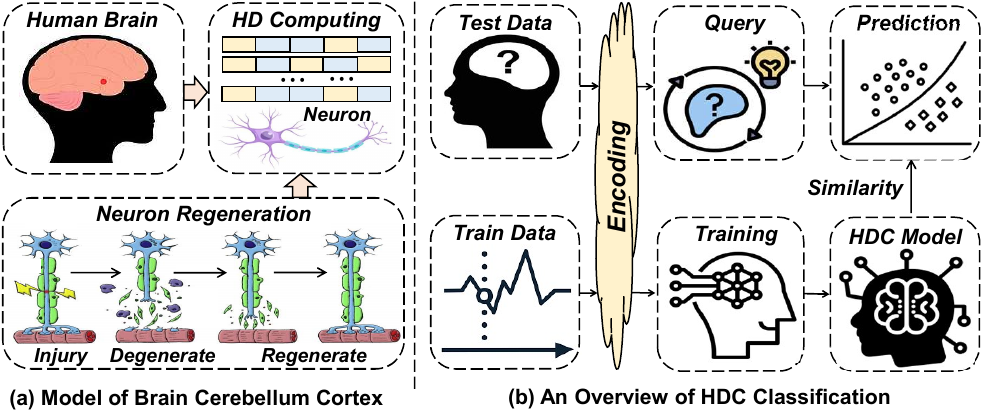}
\vspace{-7mm}
\caption{An Overview of Brain Cerebellum Cortex and HDC Classification}
\vspace{-6mm}
\label{fig: intro}  
\end{figure}
\section{Approach and Uniqueness}

\subsection{Hyperdimensional Computing}
As demonstrated in Fig.\ref{fig: intro}(b), HDC starts with encoding data points into high-dimensional space with encoding methods based on the data type. 
We then bundle encoded data by scaling a proper weight to each of them depending on how much new information they bring to class hypervectors. 
In particular, for a new encoded training sample $\mathcal H$, we update the model base on its cosine similarities with all class hypervectors, i.e. $\delta_l=\frac{ {\mathcal H} \cdot {\mathcal C_l}}{\| {\mathcal H}\|\cdot \| {\mathcal C_l}\|}$, where $ {\mathcal H} \cdot {\mathcal C_l}$ is the dot product between ${\mathcal H}$ and a class hypervector ${\mathcal C_l}$. 
For the inference phase of HDC, we encode inference data with the same encoder utilized in training to generate a query hypervector, and classify it to the class where it achieves the highest cosine similarity score.


\subsection{Identification of Undesired Dimensions}

\subsubsection{Insignificant Dimensions}
\begin{figure}[!t]
\vspace{-1mm}
\centering
\includegraphics[width=\linewidth]{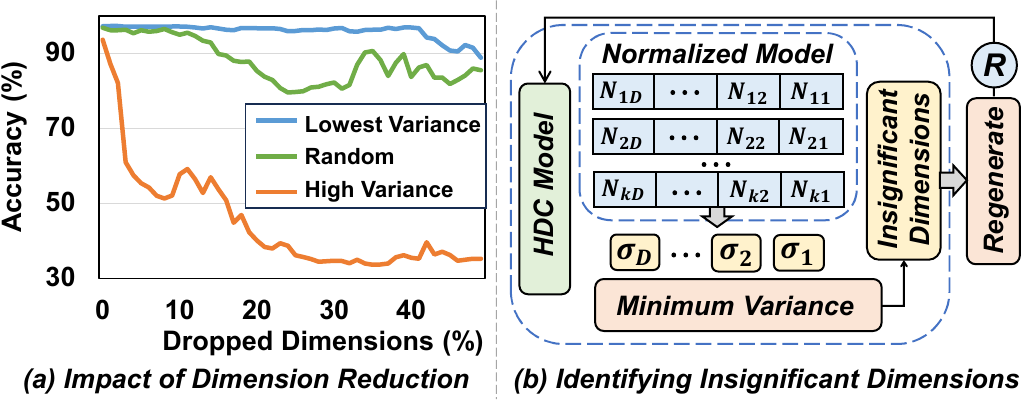}
\vspace{-7mm}
\caption{Impact and Identification of Insignificant Dimensions}
\vspace{-2mm}
\label{fig: insignificant}  
\end{figure}

HDCs represent each class with a class hypervector encoding patterns of that class. 
An effective classifier achieves the desired accuracy by a strong capability to distinguish patterns so that, in the inference phase, query vectors can have very differentiated cosine similarities to each class. 
In contrast, dimensions with similar values indicate they store common information across classes and hence play minimal roles in the classification. 
To eliminate insignificant dimensions, we calculate the variance of each dimension over all classes to measure the dispersion of that dimension. 
In particular, dimensions with minimal variances are considered insignificant. 
Fig. \ref{fig: insignificant} shows the impacts of  dimension reduction on classification accuracy. In our evaluation, dropping low variance dimensions has almost no impact on the accuracy while dropping higher variance dimensions results in a significant accuracy drop. 
We then identify and regenerate the $\mathcal R$ portion of dimensions with the lowest variance to drop, where $\mathcal R$ is the  regeneration rate. 
Compared to state-of-the-art (SOTA) HDCs, our work~\cite{wang2023late} demonstrates comparable accuracy using $8.0\times$ dimensionalities, effectively reducing the computational and memory resources needed, and delivers $1.85\times$ faster training and $15.29\times$ faster inference. 


\subsubsection{Misleading Dimensions}
\begin{figure}[!t]
\centering
\includegraphics[width=\linewidth]{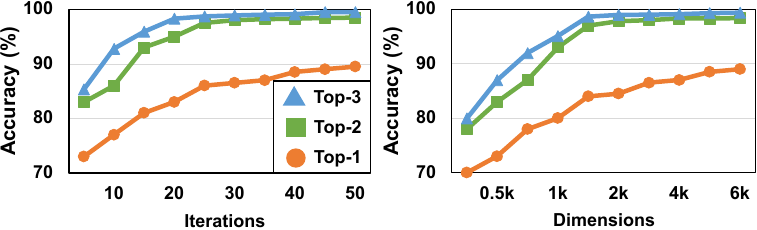}
\vspace{-7mm}
\caption{Top-1, Top-2, Top-3 Accuracy of SOTA HDCs}
\vspace{-3mm}
\label{fig: misleading}  
\end{figure}

\begin{figure}[!t]
\centering
\includegraphics[width=\linewidth]{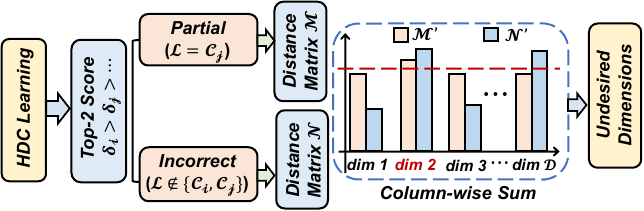}
\vspace{-7mm}
\caption{Identifying Misleading Dimensions}
\vspace{-5mm}
\label{fig: disthd}  
\end{figure}

As shown in Fig. \ref{fig: misleading}, 
SOTA HDCs provide considerably higher accuracy and faster convergence for $\textit{top-2 classification}$ than top-1 classification. We define a top-\textit{k} classification for a given data point as \textit{correct} if the true label is one of the \textit{k} most similar classes selected. 
Additionally, the accuracy difference between top-2-classification and top-3 classification is noticeably smaller than that between the top-1 classification and top-2-classification. 
Based on this, as shown in Fig. \ref{fig: disthd}, for each mispredicted data sample, we calculate the distance between the data sample and the two class hypervectors where the sample achieves the highest cosine similarities. 
We then identify and regenerate the misleading dimensions by selecting those closest to the incorrect label and farthest from the correct label to enhance model performance. 
Our work~\cite{wang2023disthd} achieves on average $2.12\%$ higher accuracy than SOTA HDCs while reducing the required dimensionalities by $8.0\times$.
It delivers $5.97\times$ faster training than SOTA DNNs and $8.09\times$ faster inference than SOTA HDCs.

\subsubsection{Biased Dimensions}
\begin{figure}[!t]
\centering
\includegraphics[width=0.9\linewidth]{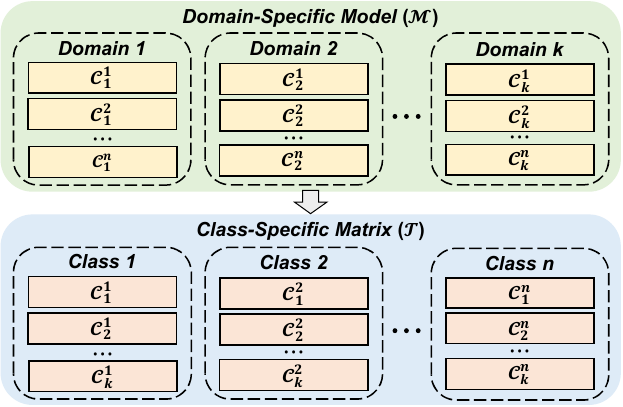}
\vspace{-3mm}
\caption{Identifying Domain-Variant Dimensions}
\vspace{-5mm}
\label{fig: domino}  
\end{figure}
A critical issue across data-driven ML approaches is \textit{distribution shift}. 
It occurs when a model is deployed on a data distribution different from what it was trained on, and can substantially degrade model performance.
As detailed in Fig. \ref{fig: domino}, to eliminate domain-variant dimensions, we first construct domain-specific hyperdimensional models and utilize these models to form class-specific matrices. 
We then calculate the variance of each dimension for each class-specific matrix. 
Dimensions with large variance indicate that, for the same class, they store very differentiated information, and are hence considered domain-variant. 
We sum up the variance vector of each class-specific matrix and filter out the top $\mathcal R$ portion of dimensions with the highest variance.
Our work~\cite{wang2023domino} provides on average $2.04\%$ higher accuracy than DNN-based domain generalization approaches, and delivers $7.83\times$ faster training and $26.94\times$ faster inference.
It also exhibits notably better performance when learning from partially labeled data and highly imbalanced data, and provides $10.93\times$ higher robustness against hardware noises than SOTA DNNs.

\subsection{Dimension Regeneration}
We utilize an encoding method inspired by the Radial Basis Function (RBF)~\cite{rahimi2007random} for  dimension regeneration. 
For an input vector in original space $\mathcal F = \{f_1, f_2, \ldots, f_n\}(f_i\in \mathds{R}
)$, we generate the corresponding hypervector $\mathcal H=\{h_1, h_2, \ldots, h_{\mathcal{D}}\} (0 \leq h_i \leq 1, h_i \in  \mathds{R})$ with ${\mathcal D} 
 (\mathcal D\gg n)$ dimensions by calculating a dot product of $\mathcal F$ with a randomly generated vector as $h_i = \cos(\mathcal B_i\cdot \mathcal F+c)\times \sin (\mathcal B_i\cdot \mathcal F)$, where $\mathcal B_i = \{b_1, b_2, \ldots, b_n\}$ is a randomly generated base vector with $b_i\sim \textit{Gaussian}(\mu =0,\sigma =1) \textrm{ and } c\sim \textit{Uniform}[0, 2\pi].$ 
We replace each base vector of the selected dimensions in the encoding module with another randomly generated vector from the Gaussian distribution and retrain the model. 

\section{Results and Contributions}
Dynamic HDC learning framework with brain-like neural adaptation has led to \textbf{three first-authored papers} in premier EDA conferences such as DAC~\cite{wang2023disthd, wang2023late} and ICCAD~\cite{wang2023domino}. It has been applied in multiple real-world applications, including cyber-security~\cite{wang2023late} and multi-sensor human activity recognition~\cite{wang2023domino}, and has been proven to significantly outperform SOTA HDCs~\cite{rahimi2016robust} in terms of training and inference efficiency.

\newpage
\bibliographystyle{unsrt}
\bibliography{ref}
\end{document}